\documentclass{article}

\usepackage[numbers]{natbib}

\usepackage {arxiv}

\usepackage{cite}
\usepackage{floatrow}
\usepackage{amsmath,amssymb,amsfonts}
\usepackage{algorithm}
\usepackage{algorithmic}
\usepackage{textcomp}
\usepackage{xcolor}
\usepackage{float}
\usepackage{booktabs}
\usepackage{enumitem}
\usepackage{setspace}
\usepackage[]{graphicx}
\usepackage{amsmath}
\usepackage{amssymb}
\usepackage{epsfig}
\usepackage{epstopdf}
\usepackage{subfigure}
\usepackage{comment}
\usepackage{tabularx}
\usepackage{threeparttable}
\usepackage{adjustbox}
\usepackage{soul}
\usepackage{graphicx}
\usepackage{subcaption}
\usepackage{hyperref} 

\def\BibTeX{{\rm B\kern-.05em{\sc i\kern-.025em b}\kern-.08em
    T\kern-.1667em\lower.7ex\hbox{E}\kern-.125emX}}

\title{\LARGE \bf
Causal Explainability of Machine Learning in Heart Failure Prediction from Electronic Health Records }

\author{Yina Hou, Shourav B. Rabbani, and Manar D. Samad\\
Department of Computer Science\\
 Tennessee State University\\
 Nashville, TN, USA\\
\texttt{msamad@tnstate.edu} \\
\and
 \bf {Liang Hong}\\
 Department of Electrical and Computer Engineering\\
Tennessee State University\\
 Nashville, TN, USA\\
\and
 \bf {Norou Diawara}\\
 Department of Mathematics and Statistics\\
Old Dominion University\\
 Norfolk, VA, USA\\
 }

\begin{document}

\maketitle

\begin{abstract}
The importance of clinical variables in the prognosis of the disease is explained using statistical correlation or machine learning (ML). However, the predictive importance of these variables may not represent their causal relationships with diseases. This paper uses clinical variables from a heart failure (HF) patient cohort to investigate the causal explainability of important variables obtained in statistical and ML contexts. Due to inherent regression modeling, popular causal discovery methods strictly assume that the cause and effect variables are numerical and continuous. This paper proposes a new computational framework to enable causal structure discovery (CSD) and score the causal strength of mixed-type (categorical, numerical, binary) clinical variables for binary disease outcomes. In HF classification, we investigate the association between the importance rank order of three feature types: correlated features, features important for ML predictions, and causal features. Our results demonstrate that CSD modeling for nonlinear causal relationships is more meaningful than its linear counterparts. Feature importance obtained from nonlinear classifiers (e.g., gradient-boosting trees) strongly correlates with the causal strength of variables without differentiating cause and effect variables. Correlated variables can be causal for HF but are rarely found as effect variables. These results can be used to add the causal explanation of variables important for ML-based prediction modeling.

\end{abstract}



\section{Introduction}

Electronic health records (EHR) are integral to modern healthcare systems, providing patient care management using digital versions of patients’ medical history~\citep{yang2023machine}. Many retrospective medical studies have been designed using EHR medical history data, including vital signs, laboratory results, diagnostic codes, prescribed medications, and quantitative measurements of anatomical and physiological features~\citep{pendergrass2019using, Samad2019Jaac}. Advanced machine learning has significantly improved personalized and patient-specific diagnosis and treatment planning~\citep{Samad2018_AHA}. In addition to patient risk stratification, machine learning models can provide scores to explain the importance of individual variables in predictive modeling~\citep{Samad2018_TOF}. In contrast, medical studies that are highly dependent on statistical correlation are often challenged by the well-known saying that 'correlation does not imply causation.' Likewise, machine learning models discover relationships between predictor variables and outcomes, which may not imply a causal relationship. It is unknown whether the variables important for machine learning are causally responsible for the outcomes.

The cause-and-effect relationship can be discovered in causal structures using directed acyclic graphs (DAG). Causal structure discovery (CSD) methods usually treat cause and effect variables of the same type. That is, all variables could be continuous or discrete. Although CSD for continuous features has been extensively studied, similar methods for discrete data remain limited, as reported in~\citep{cai2018causal} and \citep{bubnova2021mixbn}. Such models fail on binary or discrete variables, such as smoking status, and in identifying the causal structure for binary disease outcomes. Most health science studies investigate the cause-and-effect relationship between a pair of continuous variables~\citep{uchida2022medical, kotoku2020causal, naik2024applying}, without considering causal variables for a binary diagnosis of disease. Specifically, the causal relationship between a binary outcome of a disease and continuous clinical variables is crucial for discovering disease-specific causal markers and identifying variables important for disease classification. This paper proposes a computational framework that bridges the gap between machine learning modeling and causal discovery, providing an explainable analysis of cause and effect in binary disease classification. The explainable analysis introduces a means to score the causal strength and rank order causal features accordingly. Our proposed approach investigates whether the most important features for machine learning predictions are those with the highest causal strengths. The concordance between feature importance and causal strengths may reveal a novel approach to explaining decisions made by opaque machine learning models.

The remainder of the paper is organized as follows. Section \ref{background} presents the state-of-the-art causal discovery methods and their applications to health science and electronic health records. Section \ref{methods} provides a general framework for causal discovery, the proposed method for relating machine learning to causal discovery, and experimental scenarios. Section \ref{results} presents the results of the experimental scenarios. Section \ref{discussion} summarizes the results with a discussion. We conclude the paper in Section \ref{conclusions}. 

\section{Background} \label{background}

The variety of causal discovery solutions proposed in the literature can be broadly categorized into three: constraint-based, score-based, and function-based. Constraint-based algorithms such as Fast Causal Inference (FCI) \citep{spirtes2013causal} and Peter-Clark (PC) (\citep{pearl2009causality} and \citep{shen2020}) rely on conditional independence tests to identify the causal network \citep{triantafillou2014learning}.      Although suitable for high-dimensional data, these methods are computationally expensive and sensitive to noisy data or small sample sizes~\citep{verny2017learning, jia2022causal}. Moreover, constraint-based methods are unsuitable for estimating the causal strengths of individual variables, limiting the explainability of data-driven modeling~\citep{upadhyaya2023scalable}. Score-based methods, such as Greedy Equivalence Search (GES), use predefined score functions such as Akaike's Information Criterion (AIC) or Bayesian Information Criterion (BIC) to find the causal graph with the highest score among all possible DAGs\citep{aragam2024greedy}. Score-based approaches are computationally faster than their constraint-based counterparts~\citep{upadhyaya2023scalable}. However, some score-based approaches (e.g., GES) focus on achieving structural accuracy without explicitly estimating the causal strength between variables~\citep{ zhu2024hybrid}. Score-based methods often use local heuristics to navigate the search space because enforcing acyclicity is a complex combinatorial problem~\citep{zhang2021gcastle}. The authors of Non-combinatorial Optimization via Trace Exponential and Augmented Lagrangian for Structure (NOTEARS)~\citep{zheng2018dags, zheng2020learning} have overcome this issue of the score-based method by replacing discrete DAG constraints with a smooth equality constraint, turning causal discovery into a continuous optimization problem~\citep{yang2024federated}. However, the performance of similar score-based methods can degrade in the presence of non-linear relationships~\citep{li2024nonlinear}. These linear models can be too simplified for real-world medical data that exhibit complex and non-linear relationships~ \citep{shen2020novel}. Researchers have extended causal discovery frameworks to nonlinear models. For example, NOTEARS is upgraded by incorporating a multilayer perceptron to approximate the underlying non-linear causal relationships in the data ~\citep{zheng2020learning}.

Function-based methods, on the other hand, such as Linear Non-Gaussian Acyclic Model (LiNGAM)~\citep{shimizu2006linear} and its variants like DirectLiNGAM ~\citep{shimizu2014direct} make several assumptions to obtain causal structures.  The baseline LiNGAM assumes linearity in causal relationships with non-Gaussian data distributions by leveraging statistical independence properties, such as independent component analysis (ICA). Its ability to handle non-Gaussian noise has made LiNGAM successful in many real-world data problems, including EHR data analysis \citep{uchida2022medical}, where non-Gaussian noise distribution often arises from measurement errors or patient variability. However, the LiNGAM model is likely to converge to local minima, affecting the accuracy of the DAG model~\citep{cai2024learning}. To overcome this problem, DirectLiNGAM is proposed by~\citep{shimizu2011directlingam}. DirectLiNGAM uses linear regressions instead of ICA to determine causal order by testing mutual independence between the independent variable and the residuals. Uchida.~\citep{uchida2022medical} use the DirectLiNGAM model to classify an ordinal variable that represents the severity of fatty liver disease and shows the causal strength of individual features to the severity of the disease outcome. However, the authors have excluded categorical or discrete variables, stating the limitation of the DirectLiNGAM model in handling such data types.

Due to the constituent regression models, the application of these popular causal discovery methods is limited to continuous features~\citep{glymour2019review, mesner2020non, zanga2022survey}. These models fail on nonordinal discrete features, lacking natural ordering~\citep{cai2018causal}. To address this issue, hybrid models such as Hybrid Causal Discovery in Mixed Type Data (HCM) ~\citep{li2022hybrid}, NOTEARS-M~\citep{zhao2024notears} for mixed data types, and PC-NOTEARS ~\citep{zhu2024hybrid} have been introduced to handle causal discovery for mixed data types. However, NOTEARS-M is restricted to linear causal relationships~\citep{li2022hybrid, zhu2024hybrid} while none of these methods considers causal strengths between variables when evaluating causal relationships.

The literature gaps highlight the need to extend the causal strength scoring to nonlinear relationships and mixed variables, especially in explaining causal relationships in machine learning outcomes. Specifically, the relationship between the importance of variables in machine learning and the causality strengths of these variables has not been well explored in the literature. Recently, Rashid et al. \citep{rashid2023causal} have used DirectLiNGAM first to obtain causal variables affecting two disease outcome variables. However, the model using selective causal variables has resulted in worse disease classification performance than the model using all variables.  Their results do not explain whether the drop in classification performance is due to the simplified linearity assumption of DirectLiNGAM or whether such models are ineffective in handling discrete outcome variables. 

The paper aims to address these gaps through the following contributions. First, the causal discovery between continuous predictor features and discrete disease labels is enabled by transforming discrete labels into continuous likelihood scores for individual samples. Second, the causal explainability of machine learning prediction is investigated by comparing variable importance scores with scores of causal strengths.

\section {Methodology} \label{methods}
This section presents a generic framework for the causal discovery model, the proposed computational framework, and the experimental steps, including data curation methods, for electronic health records.

\subsection{Causal structure discovery}
The primary objective of causal structure discovery (CSD) is to identify a directed acyclic graph (DAG) using an adjacency feature matrix \( A \in \mathbb{R}^{N \times N} \) built from a tabular data set \( X \in \mathbb{R}^{B \times N} \) with \( B \) samples and \( N \) variables. The DAG encodes the causal relationships between variables and is optimized to best explain the observed dependencies and variations in the data. The learning of the DAG can be framed as a minimization problem as follows:
\[
\arg\min_A f(X, A) \quad \text{where} \quad A_{i,j} = 0 \text{ when } i = j.
\]
Here, the loss function \( f(X, A) \) is minimized with a constraint that ensures the graph is acyclic. Each entry \( A_{i,j} \) represents the strength of the causal relationship between features \( x_i \) (cause) and \( x_j \) (effect). 

This paper uses two representative causal discovery models. First, DirectLiNGAM is a linear function-based method that provides the most explainable scoring for causal strength. A positive score indicates that the increase of one variable causes the effect variable to increase, and vice versa. Second, NOTEARS-MLP is the non-linear version of the score-based method, NOTEARS. These two methods for discovering causal structures are explained in subsequent sections.

\subsection{DirectLiNGAM}
The baseline LiNGAM model produces a suboptimal solution when the number of variables increases. This is overcome in DirectLiNGAM, which estimates causal relationships using a series of least-square regressions along with independence tests \citep{cai2024learning}. During each iteration, DirectLiNGAM uses a linear regressor to predict each feature using a linear combination of the remaining features, as follows. 
\begin{equation}
\mathbf{X} = \mathbf{\beta} \mathbf{X} + \mathbf{G} \mathbf{z} + \mathbf{e}.
\label{eq:lingam_regressor}
\end{equation}

Here, \(\mathbf{z}\) is the unobserved confounding variable, the matrix \(\mathbf{G}\) explains how the unobserved confounding variables affect the observed variables and \(\mathbf{e}\) is the noise. In the DirectLiNGAM model, the effect of confounding variables is ignored as \( \mathbf{z=0}\). The variable with the most independent residual, determined by mutual information, is selected as the causal variable. In other words, the residual \( e_j \) of the variable \( j \) that produces the lowest total mutual information with all other variables is considered causal. This is achieved by minimizing the total mutual information (MI) between the residuals of a candidate variable and all other variables, formalized as:
\begin{equation}
\min \sum_{i \neq j} MI(e_j, X_i)
\label{eg:lingam_mi}
\end{equation}
The two steps in Equations \ref{eq:lingam_regressor} and \ref{eg:lingam_mi} are repeated for each of the \( N-1 \) variables to find causal relationships among all features. When a causal variable is identified, it is excluded before iteratively analyzing the remaining variables until all causal relationships are discovered. For example, for a pair of features P and Q, DirectLiNGAM will consider both possible directions: (1) \( P \to Q \) and (2) \( Q \to J \) and select the one with the lowest MI. In the presence of multiple EHR features, DirectLiNGAM progressively removes causal features with the lowest MI, refining the discovery of the correct causal structures.

\subsection{NOTEARS-MLP}
NOTEARS-MLP\citep{zheng2020learning} is a nonlinear extension of the baseline NOTEARS \citep{zheng2018dags} using multiple multilayer perceptron (MLP) regressors. Each MLP regressor model $M_j$ estimates a target feature $X_j$, taking N variables as input. The corresponding MLP weight of $X_j$ is set to zero to avoid its contribution to reconstruction. The first layer of $M_j$ uses a weight matrix $\theta^1_j$ of size \((N, D)\) to map $N$ variables to a $D$-dimensional embedding. The estimated value $\hat{X}_j$ is obtained using the following MLP. 
\begin{equation}
\label{eq:MLP}
\hat{X}_j = M_j (X; \theta_j ) = \sigma\left(\sigma\left(\dots\sigma(X \theta_j^1) \theta_j^2 \dots\right) \theta_j^H \right),
\end{equation}

where, \( \theta_j^l \in \mathbb{R}^{d_l \times d_{l+1}} \) represents the weights of the \( l \)-th layer of the \( j \)-th MLP, and \( \sigma \) is the nonlinear activation function. H is the number of layers in the MLP. NOTEARS-MLP computes an adjacency matrix $W$ from the first layer of each MLP regressor to infer causal relationships. The weight matrices \(\theta_j^1 \in \mathbb{R}^{N \times D}\) from the first layers of the $N$ MLPs are aggregated as a tensor of shape \(N \times N \times D\). Applying the L2-norm along the last dimension produces the adjacency matrix $W(\theta) \in \mathbb{R}^{N \times N}$. This process is summarized in Equation~\ref{eq:weight} as:

\textbf{\begin{figure*}
    \centering
    \includegraphics[width=1.0\linewidth]{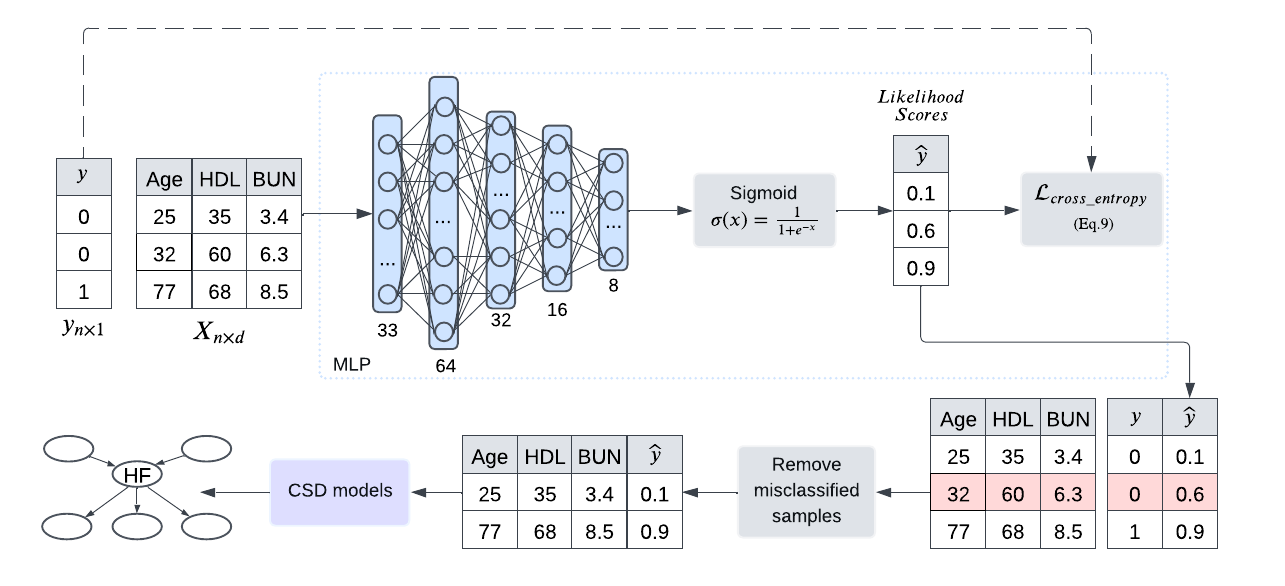}
    \vspace{-10pt}
    \caption{Proposed computational framework for causal structure discovery between continuous variables and binary disease outcome.} 
    \label{fig:diagram}
\end{figure*}}
\begin{equation}
\label{eq:weight}
W_{kj} = \| T_{k, j} \|_2, \text{ where } T = [\theta^1_1, \ldots, \theta^1_N].
\end{equation}

To ensure that the learned graph remains a DAG, NOTEARS-MLP uses the penalty term proposed in \citep{zheng2018dags}. This term relies on k walks, where a non-zero trace ($ tr(W^k) \neq 0$) indicates the presence of cycles. However, when $W$ contains positive and negative entries, cancellation effects can drive the trace toward zero. To mitigate this, the Hadamard product ($W \odot W$)  is applied element-wise to remove negative entries. The infinite sum of the k-walk is then approximated using its Taylor series expansion, resulting in $e^{W \odot W}$. However, this approximation includes the contribution of 0-walk, which results in \(\text{tr}(e^{W^0}) = N\). To correct for this, \(N\) is subtracted from the closed form. This process is shown in Equation~\ref{eq:trace}.
\begin{equation}
\label{eq:trace}
h(W) = \text{tr}(e^{(W \odot W)}) - N.
\end{equation}
Here, $h(W)$ provides a way to detect cycles numerically. However, the function is non-convex and leads to suboptimal solutions~\citep{zheng2018dags}. This is fixed using the augmented Lagrangian, which penalizes small changes in $W$ that lead to bad solutions. This updates Equation~\ref{eq:trace} to Equation~\ref{eq:loss_dag}.

\begin{equation}
\label{eq:loss_dag}
L_{\text{DAG}} = \alpha h(W(\theta)) + \frac{\rho}{2} |h(W)|^2 .
\end{equation} 
The hyperparameters $\alpha$ and $\rho$ regulate the acyclicity penalty strength. Since the latter penalizes changes more heavily, its strength is halved.
The loss function to optimize all MLP regressors includes three components: mean-squared loss of \( N \) regressors, \( L_1 \) regularization to introduce sparsity, and the penalty to enforce acyclicity ($L_{DAG})$, as demonstrated in Equation~\ref{eq:loss}.
\begin{equation}
\label{eq:loss}
\min_{\theta} F(\theta) =
\frac{1}{N} \sum_{j=1}^{N} \|X_j - \hat{X}_j \|_F^2 + \lambda\|\theta^1\|_1 + L_{\text{DAG}}.
\end{equation}
Here, \( \lambda \) is a hyperparameter to regulate the sparsity in the first layer. This approach enables NOTEARS-MLP to learn nonlinear causal relationships while ensuring the graph remains a Directed Acyclic Graph (DAG). Unlike DirectLiNGAM, which tests specific causal directions, NOTEARS-MLP employs a continuous optimization approach with neural networks to learn causal structures and capture nonlinear relationships. However, both methods rely on regression-based techniques and struggle with discrete features, such as binary or categorical outcomes. We address this limitation in the next section.
\subsection{Proposed Computational Framework}
We propose several computational steps in line with our contributions, as presented in Figure~\ref{fig:diagram}. Discrete labels to continuous features: Causal discovery methods that can provide causal strengths require both cause and effect variables to be continuous. In contrast, classification labels in machine learning are discrete or binary. The cause-and-effect relationship between predictor variables and a disease diagnosis (binary label), such as heart failure, can share valuable clinical insights. We utilize a multi-layer perceptron (MLP) to transform a discrete binary label into continuous-valued likelihood scores, enabling the discovery of causal structures between a disease and predictor variables. An MLP yields likelihood scores for a disease label given input features using a non-linear activation function.
\begin{equation}
\hat{y} = \sigma(\text{MLP}(X)).
\end{equation}
Here, \( \mathbf{X} \) is the input, \( \text{MLP}(X) \) represents the MLP outputs as logit scalar, and \( \sigma(\cdot) \) applies the sigmoid function to the MLP output to produce the likelihood scores \( \hat{y} \). The MLP training updates the likelihood scores following the actual class labels, which is terminated early to prevent overfitting. Continuous-valued likelihood scores, representing the non-linear relationship between the predictors and disease outcome, are used as a proxy for discrete class labels in subsequent causal discovery.
\begin{figure*}[t]
    \centering
    \subfigure[Binary class labels]{\includegraphics[width=0.49\textwidth]{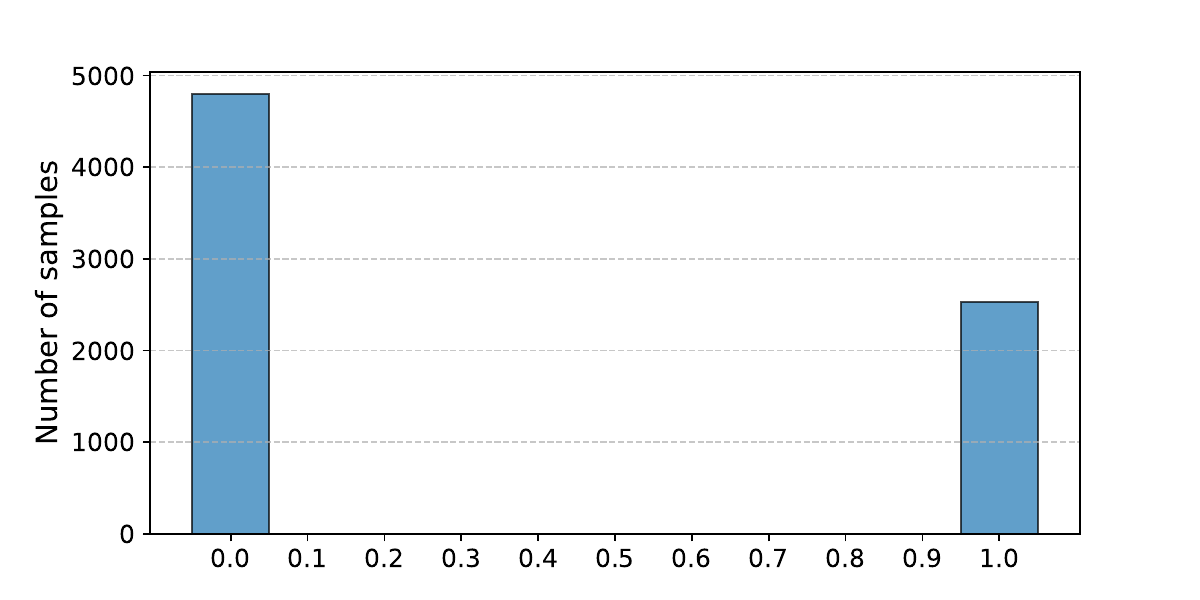}}
    \hfill
    \subfigure[Continuous likelihood scores]{
    \includegraphics[width=0.49\textwidth]{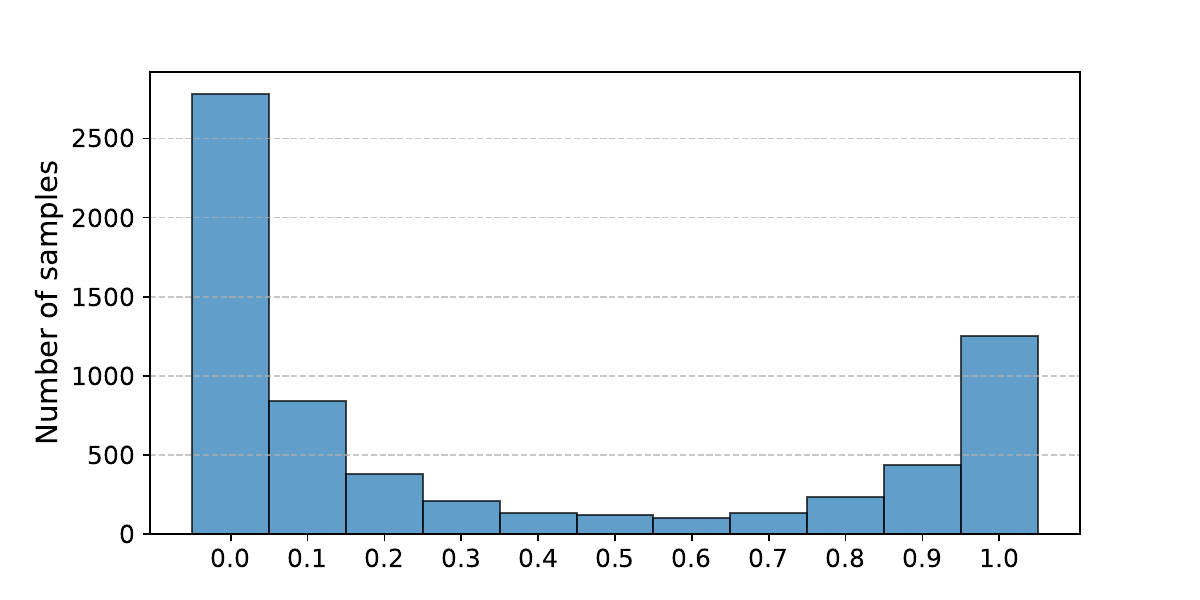}}
   \vspace{-10pt}
    \caption{Histogram of binary \{0, 1\} class labels and continuous likelihood scores between 0 and 1.}
    \label{fig:distribution}
\end{figure*}
\subsection{Explain feature importance using causal strengths. }
The most accurate machine learning models are typically complex, with outcomes that can be difficult to explain. Many research domains, including health science, rely on feature importance scores to explain the predictive analysis of the model~\citep{wang2024recent}. The gradient-boosted trees (GBT) classifier is one of the best-performing machine learning methods for tabular data, which additionally provides feature importance scores associated with predicting the class label. Feature importance scores are used to rank features. Similarly, scores obtained from the adjacency matrix of a causal discovery graph can represent the causal strengths of individual features, allowing them to be ranked in order. Previous studies~\citep{huang2021benchmarking, uchida2022medical} have used the graph adjacency matrix to obtain causal strengths in cause-effect analyses between continuous features. However, these studies have not been conducted on binary outcome prediction, where the direction of causality in the DAG structure can be interpreted in two ways. First, predictor variables (causal variables) can cause the disease. Second, predictor variables (also known as effect variables) are caused by the disease. The correlation between the rank order of the important variables for machine learning and that of the causal variables can provide valuable insights for health data analytics.

\subsection{Experimental setting }
This section presents the experimental steps, cases, and evaluation methods.
\subsubsection{Discrete to continuous likelihood scores} The first step of our framework involves transforming discrete outcome labels into continuous likelihood scores using a multilayer perceptron (MLP). The MLP architecture consists of four hidden layers with decreasing neuron counts: 64, 32, 16, and 8, and a single output neuron for binary classification. The MLP is trained on the dataset using a class-weighted binary cross-entropy loss to handle any class imbalance in the data and is defined as follows.
\begin{equation}
\mathcal{L} = -\mathbb{E} \left[ w_1 \cdot y \log(\hat{y}) + w_0 \cdot (1 - y) \log(1 - \hat{y}) \right],
\end{equation}
where, $w_0$ and $w_1$ are the class weights for false and positive HF samples. $y$ and $\hat{y}$ are the true and predicted class labels. The model training continues until the model achieves 90\% accuracy, striking a balance between accuracy and overfitting. Following an early stop of MLP training, misclassified samples are excluded and correctly classified samples with their likelihood scores are used in subsequent causal discovery. The likelihood scores of correctly classified samples are assumed to be the correct proxy of their true class labels. Class likelihood scores are concatenated with predictor variables to discover causal structures between all possible variables and between class and predictor variables.

\textbf{\begin{figure*} [t]
    \centering
    \includegraphics[width=1.0\linewidth]{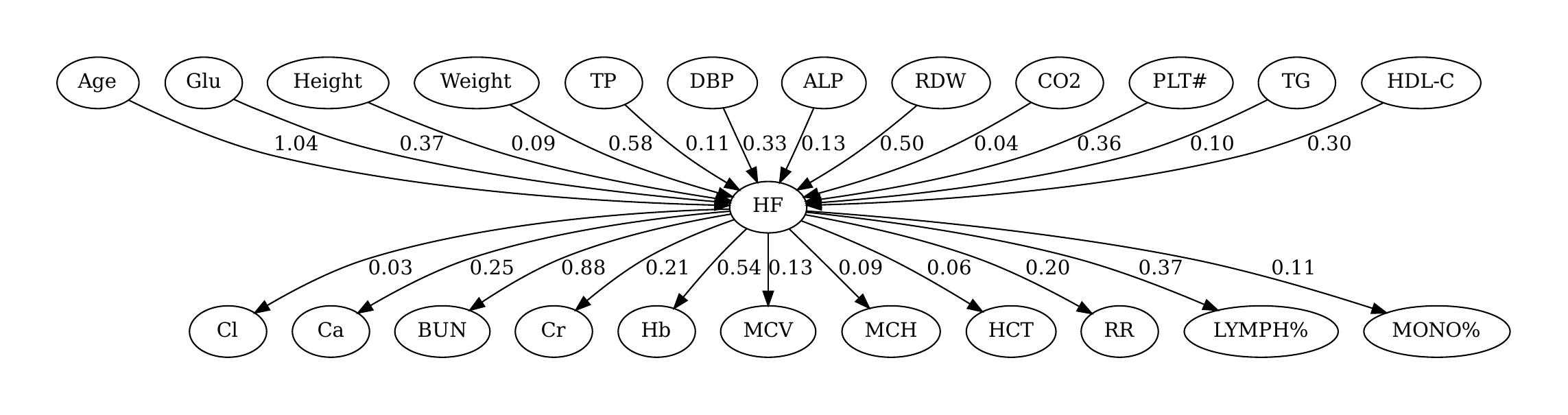}
    \vspace{-20pt}
    \caption{Direct Acyclic Graph (DAG) generated by the NOTEARS-MLP model demonstrating the cause (X$\rightarrow$ HF) and effect (HF $\rightarrow$ X) relationship between HF diagnosis and EHR features.}
    \label{fig:NOTEARS_MLP_dag}
\end{figure*}}

\textbf{\begin{figure*}
    \centering
    \includegraphics[width=1.0\linewidth]{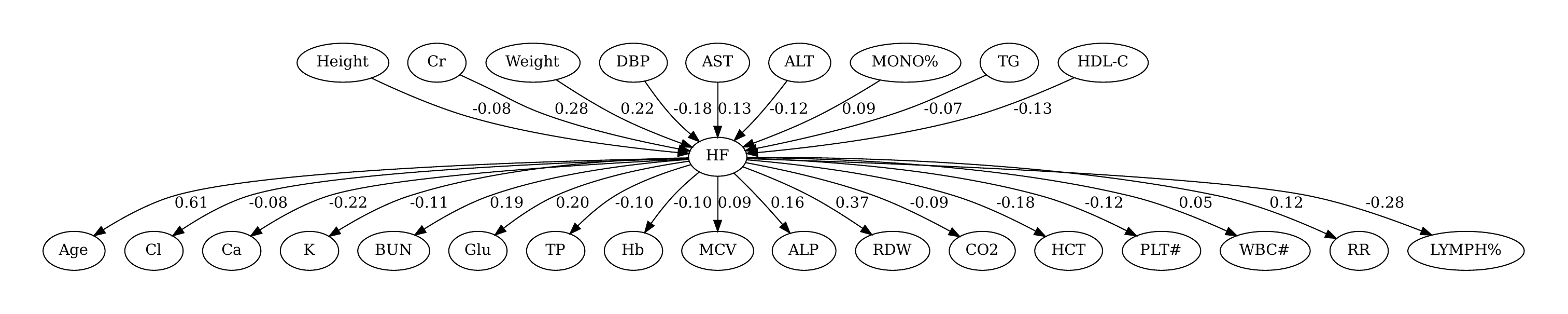}
     \vspace{-20pt}
    \caption{Direct Acyclic Graph (DAG) generated by the DirectLiNGAM model demonstrating the cause (X$\rightarrow$ HF) and effect (HF $\rightarrow$ X) relationship between HF diagnosis and EHR features.}
    \label{fig:LiNGAM_dag}
\end{figure*}}

\begin{table*}[t]
\caption{Pearson correlation coefficients between feature values and HF likelihood scores. Statistical significance (p-value $< 0.05$) is denoted by *. } 
\label{tab:pearson_correlation_feature_likelihood_horizontal}
\scalebox{0.65}{ 
\centering
\begin{tabular}{c|ccccccccccccccccc}
\toprule
\textbf{Feature}        & Age    & BUN    & RDW    & Cr      & Glu    & Weight & ALP    & RR     & WBC\#  & MONO\% & AST    & K      & BASO\% & MCV    & EOS\%  & SBP   & HR \\
\midrule
\textbf{Positive Corr.} & 0.55*  & 0.43*  & 0.39*  & 0.29*   & 0.23*  & 0.19*  & 0.17*  & 0.12*  & 0.12*  & 0.10*  & 0.04*  & 0.02   & 0.01   & 0.01   & 0      & 0     & 0  \\
\toprule
\textbf{Feature}        & Hb     & HCT    & RBC\#  & LYMPH\% & Ca     & TP     & PLT\#  & DBP    & HDL-C  & CO\_2  & Na     & MCH    & Cl     & ALT    & Height & TG    &    \\
\midrule
\textbf{Negative Corr.} & -0.44* & -0.40* & -0.35* & -0.33*  & -0.26* & -0.22* & -0.19* & -0.18* & -0.17* & -0.13* & -0.12* & -0.11* & -0.06* & -0.03* & -0.03* & -0.02 &   \\
\bottomrule
\end{tabular}}
\end{table*}
\subsubsection{Machine learning variable importance}
In the second step, the samples correctly classified in the previous step are used to obtain feature importance scores and rank order by training a Gradient Boosting Tree (GBT) classifier. Individual feature importance scores from a 10x2 cross-validation are averaged across ten outer test data folds. The average importance scores are used to obtain the rank order of individual variables, where the most important variable has the highest rank (1). For comparison with the non-linear GBT classifier, we obtain the variable importance rank using a linear classifier model, logistic regression. 

\subsubsection{Experimental cases and evaluation}

The variable rank orders are obtained for four experimental cases: 1) causal variables, 2) effect variables, 3) important variables for machine learning, and 4) variables correlated with disease class likelihood scores. Cases 1 and 2 are obtained using a linear CSD model (DirectLiNGAM) and a non-linear CSD model (NOTEARS-MLP). Case 3 is obtained using linear (logistic regression) and non-linear (GBT) machine learning classifiers. Case 4 correlates individual variables with the likelihood score for the disease class. The correlation between rank orders is obtained using Spearman correlation coefficients. Spearman correlation is particularly suitable for ordinal values, without requiring any assumptions about data distribution. The statistical significance p-value for each correlation is reported at a significance level of $\alpha$ = 0.05.

\section {Results}  \label{results}
The results of the data processing and analysis are discussed in the following.

\subsection {Heart Failure EHR Data}
A real-world heart failure (HF) classification data set is created from the All of Us Research Program supported by the US National Institutes of Health (NIH)~\citep{allofus}. The cohort without cardiovascular disease is labeled as zero, and those with HF diagnoses are labeled as 1 for binary classification of HF using clinical variables. Our curated EHR data include 7333 patients, 2534 of whom represent the non-CVD cohort, and 4799 of whom have HF diagnoses. The curated data set has an overall missing rate of 0.39\%, which is imputed using the median values of individual variables. We select 33 variables considering their clinical relevance and the availability of data. We aim to focus on male patients, motivated by evidence from \citep{tsao2023heart}. It indicates that HF is significantly more prevalent among men aged 20 to 79 years, making this demographic critical to understanding the dynamics of HF. All experiments involving All of Us data must be conducted within the program’s secure Researcher Workbench on the cloud to ensure the privacy of All of Us participants. We provision a cloud computing environment with a 2-core CPU with 13 GB of RAM. 

\subsection{Class likelihood scores and DAG}
Figure~\ref{fig:distribution} shows the class distribution of binary HF labels and continuous likelihood scores. The Pearson correlation between variables and class likelihood scores is presented in Table \ref{tab:pearson_correlation_feature_likelihood_horizontal}. The likelihood scores of the correctly classified samples are used to obtain DAGs and causal strengths. The DAG with causal strengths using NOTEARS-MLP and DirectLiNGAM are presented in Figures \ref{fig:NOTEARS_MLP_dag} and \ref{fig:LiNGAM_dag}, respectively.   

The node variables and the directed edges between the nodes represent the causal direction of the variables as follows. \( X \rightarrow Y \) implies \( X \) causes \( Y \) while \( Y \rightarrow X \) implies \( X \) is an effect of \( Y \). A DAG presents variables causing the disease and also variable changes caused by the disease. 
The absolute values of the weight matrix show the causal strengths on the DAG edges.

A total of 23 out of 33 variables appear in the DAG of NOTEARS-MLP. Twelve of these variables appear as causal variables for HF, where age is the most causal factor, followed by body weight and serum glucose level. In contrast, changes in eleven variables are caused by HF, with blood urea nitrogen (BUN) being the most affected by HF. The directLiNGAM linear causal discovery model shows 26 variables in its DAG representation. Only nine of these variables are causal variables for HF. 

\begin{table*}[t]
\centering
\caption{Rank order of causal variables (Causal),  machine learning important variables (GBT\_Imp, LR\_Imp), and variables correlated (F\_Corr) with HF likelihood.}
\label{tab:combined_rank}
    \begin{tabular}[t]{lcccc|lcccc}
    \toprule
    \multicolumn{10}{c}{\textbf{Causal}}\\
    \midrule
    \multicolumn{5}{c|}{\textbf{NOTEARS-MLP}} & \multicolumn{5}{c}{\textbf{DirectLiNGAM}} \\
    \midrule
    Variable & Cause & GBT & LR & F  & Feature & Cause & GBT & LR & F \\
    & & Imp & Imp & Corr  & & & Imp & Imp & Corr \\
    \midrule
    Age    & 1  & 1  & 1  & 1  & Cr     & 1 & 1 & 3 & 1 \\
    RDW    & 3  & 3  & 2  & 2  & Weight & 2 & 2 & 1 & 2 \\
    Glu    & 4  & 2  & 5  & 3  & DBP    & 3 & 5 & 8 & 3 \\
    TP     & 9  & 8  & 9  & 4  & HDL-C  & 5 & 3 & 2 & 4 \\
    Weight & 2  & 4  & 3  & 5  & MONO\% & 7 & 8 & 4 & 5 \\
    PLT\#  & 5  & 5  & 4  & 5  & AST    & 4 & 7 & 5 & 6 \\
    DBP    & 6  & 10 & 11 & 7  & Height & 8 & 9 & 6 & 7 \\
    ALP    & 8  & 7  & 7  & 8  & ALT    & 6 & 6 & 7 & 7 \\
    HDL-C  & 7  & 6  & 6  & 9  & TG     & 9 & 4 & 9 & 9 \\
    $CO_2$    & 12 & 11 & 8  & 10 &        &   &   &   &   \\
    Height & 11 & 12 & 10 & 11 &        &   &   &   &   \\
    TG     & 10 & 9  & 12 & 12 &        &   &   &   &  \\
    \bottomrule
    \end{tabular}
    \end{table*}
    
    \begin{table*}[t]
    \centering
\caption{Rank order of effect variables,  machine learning important variables (GBT\_Imp, LR\_Imp), and variables correlated (F\_Corr) with HF likelihood.}
\label{tab:combined_rank}    
    \begin{tabular}[t]{lcccc|lcccc}
    \toprule
    \multicolumn{10}{c}{\textbf{Effect}}\\
    \midrule
    \multicolumn{5}{c|}{\textbf{NOTEARS-MLP}} & \multicolumn{5}{c}{\textbf{DirectLiNGAM}} \\
    \midrule
    Variable & Effect & GBT & LR & F  & Variable & Effect & GBT & LR & F \\
    & & Imp & Imp & Corr  & & & Imp & Imp & Corr \\
    \midrule
    Hb      & 2  & 1  & 6  & 1  & Age     & 1  & 1  & 1  & 1  \\
    BUN     & 1  & 2  & 1  & 2  & Hb      & 12 & 2  & 9  & 2  \\
    HCT     & 10 & 7  & 9  & 3  & BUN     & 6  & 3  & 3  & 3  \\
    LYMPH\% & 3  & 4  & 3  & 4  & HCT     & 7  & 12 & 13 & 4  \\
    Cr      & 5  & 3  & 7  & 5  & RDW     & 2  & 5  & 2  & 5  \\
    Ca      & 4  & 5  & 5  & 6  & LYMPH\% & 3  & 6  & 5  & 6  \\
    RR      & 6  & 6  & 8  & 7  & Ca      & 4  & 8  & 7  & 7  \\
    MCH     & 9  & 9  & 4  & 8  & Glu     & 5  & 4  & 8  & 8  \\
    MONO\%  & 8  & 10 & 11 & 9  & TP      & 13 & 11 & 17 & 9  \\
    Cl      & 11 & 11 & 10 & 10 & PLT\#   & 9  & 7  & 6  & 10 \\
    MCV     & 7  & 8  & 2  & 11 & ALP     & 8  & 9  & 10 & 11 \\
            &    &    &    &    & $CO_2$     & 14 & 17 & 12 & 12 \\
            &    &    &    &    & RR      & 10 & 10 & 11 & 13 \\
            &    &    &    &    & WBC\#   & 17 & 13 & 16 & 13 \\
            &    &    &    &    & Cl      & 16 & 16 & 14 & 15 \\
            &    &    &    &    & K       & 11 & 15 & 15 & 16 \\
            &    &    &    &    & MCV     & 15 & 14 & 4  & 17 \\
    \bottomrule
\end{tabular}
\end{table*}

\subsection{ML variable importance and causal variables}
The EHR variables are ranked according to individual importance scores to classify HF diagnosis using gradient-boosted trees (GBT) and logistic regression (LR) classifiers. Likewise, variables are ranked on the basis of their correlation with HF likelihood scores. The Spearman correlation between the two variable rank orders reveals a higher correlation (0.68, $p<0.05$) for GBT than for LR (0.48, $p<0.05$).

Table \ref{tab:combined_rank} shows the rank orders of the important ML variables and the causal variables. The causal variables of NOTEARS-MLP are strongly correlated (0.89, $p<0.05$) with the important variables of GBT for classification. The same strong correlation is observed for the important variables identified by LR. In particular, the correlation between the important variables of GBT and LR classifiers is 0.62 ($p<0.05$). The causal variables of DirectLiNGAM show a low correlation (0.63, $p> 0.05$) with important variables of GBT and with important variables of LR (0.55, $p> 0.05$).

\subsection{ML variable importance and effect variables}

On the other hand, the effect variables are ranked based on the causal strengths of the HF diagnosis. The Spearman correlation between the rank order of important ML variables and that of effect variables shows a high correlation (0.90, $p<0.05$)  for the GBT classifier and NOTEARS-MLP. The effect variables of NOTEARS-MLP result in a low correlation (0.61, $p>0.05$) with important variables of the LR classifier. For DirectLiNGAM, the correlation scores are 0.73 ($p<0.05$) and 0.69 ($p<0.05$) with the GBT and LR classifiers, respectively.

\subsection{Correlation versus causality}

This section investigates whether correlation implies causation. Variables correlated with the likelihoods of HF are also highly causal variables for NOTEARS-MLP (0.82, $p<0.05$) and DirectLiNGAM (0.90, $p<0.05$). However, effect variables are less likely to be correlated variables for NOTEARS-MLP (0.65, $p>0.05$) and DirectLiNGAM (0.71, $p<0.05$). Table \ref{tab:combined_rank} shows that Age, RDW, Creatinine (Cr), Glucose (Glu), and Weight are among the top correlated variables with the likelihood of HF. NOTEARS-MLP identifies these correlated variables as the top causal variables, except Cr is an effect variable. In contrast, DirectLiNGAM identifies the most correlated variables as effect variables (age, BUN, RDW, Glu), except Cr and weight are identified as causal variables. Among the top five negatively correlated variables, Hb, HCT, LYMPH\%, and Ca appear as effect variables for NOTEARS-MLP and DirectLiNGAM. Some correlated variables do not appear as causal variables. In NOTEARS-MLP, TP is correlated but not strongly causal with HF likelihoods. BUN (Blood Urea Nitrogen) is the second most correlated variable, but appears as an effect variable in both DirectLiNGAM and NOTEARS-MLP. Although showing the third most negative correlation, RBC\# does not appear in the causal and effect variable rank order in NOTEARS-MLP and DirectLiNGAM. This finding highlights the key distinction between correlation and causation, emphasizing that high correlation alone does not always imply a direct causal relationship (0.89, $p<0.05$).

\begin{table}[t]
\caption{Spearman correlation between rank orders (X and Y) of EHR variables. Statistically not significant (p-value $> 0.05$) results are denoted by NS.}
\label{tab:corr_CEIMP}
\centering
\begin{tabular}{llcccc}
\toprule
 & &  \multicolumn{2}{c}{DirectLiNGAM} & \multicolumn{2}{c}{NOTEARS-MLP} \\
\cmidrule(lr){3-4} \cmidrule(lr){5-6}
 X & Y & $\rho$ & p-value & $\rho$ & p-value \\
\midrule
  Causal & F\_Corr  &  0.90*  &  $p<0.05$  &  0.82*  &  $p<0.05$ \\
Causal &   GBT\_Imp   &  0.63  &  NS  &  0.89*  &  $p<0.05$ \\
  Causal  &  LR\_Imp  &  0.55  &  NS  &  0.82*  &  $p<0.05$ \\
  Effect  &  F\_Corr  &  0.71*  &  $p<0.05$  &  0.65*  &  $p<0.05$ \\
 Effect  &  GBT\_Imp  &  0.73*  &  $p<0.05$  &  0.90*  &  $p<0.05$ \\
  Effect  &  LR\_Imp  &  0.69*  &  $p<0.05$  &  0.61  &  NS \\
\bottomrule
\end{tabular}
\end{table}

\section{Discussion} \label{discussion}

The summary of our experimental results is as follows. First, a non-linear causal structure discovery (e.g., NOTEARS-MLP) is more accurate and reliable than its linear counterparts (e.g., DirectLiNGAM). This superiority stems from their ability to capture non-linear relationships, addressing the limitations of linear methods. Although a variant of the NOTEARS method uses logistic regression to handle discrete variables, it remains restricted to linear causal relationships~\citep{zhao2024notears}. In contrast, our method employs an MLP, enabling the discovery of more complex, non-linear causal patterns. This statement may be validated based on standard medical knowledge that affirms age as the most causal factor in cardiovascular disorders, including HF~\citep{li2020targeting}. NOTEARS-MLP correctly identifies age as the most causal factor for HF, including weight. In contrast, age and weight are identified as effects of HF by the linear DirectLiNGAM model. Second, important variables derived from a GBT classifier can serve as a proxy for identifying both causal and effect variables obtained via NOTEARS-MLP. However, the important variables in the ML classification do not distinguish between the cause and effect variables. Third, when the superior NOTEARS-MLP is considered, the correlated variables do not imply effect variables ($\rho$ = 0.65, ($p<0.05$)), but may partially imply causal variables ($\rho$ = 0.82 ($p<0.05$)). Therefore, separating correlated variables that are not causal is a critical task. Fourth, a negative correlation may weakly imply effect variables instead of causal variables.

\section{Conclusions} \label{conclusions}
This paper presents a novel causal discovery method that can handle both categorical and numerical variables. 
The model supports mixed-type data, addressing a key limitation of existing methods that rely on continuous data assumptions. In the real-world HF dataset with binary outcomes, the proposed framework shows a strong correlation between feature importance from nonlinear machine learning models and their estimated causal strength.  Although features with high statistical correlation contribute to disease outcome, they are infrequently identified as effects in the causal structure. These results reveal the value of non-linear causal modeling in explaining predictive variables and highlight the importance of integrating causal reasoning into ML-based prognosis modeling.

\bibliographystyle{elsarticle-num}
\bibliography{mybib}

\end{document}